\documentclass[letterpaper, 10 pt, conference]{ieeeconf}  

\IEEEoverridecommandlockouts                              

\usepackage{booktabs}       
\usepackage{amsfonts}       
\usepackage{nicefrac}       
\usepackage{pifont}
\usepackage{xcolor}         
\usepackage{multirow}
\usepackage{microtype}      
\usepackage{graphicx} 
\usepackage{amsmath}
\usepackage{amssymb}
\usepackage{wrapfig}
\usepackage{colortbl}
\usepackage{url}
\usepackage[colorlinks,citecolor=ForestGreen]{hyperref}

\def\check{$\checkmark$}

\newcommand{\myParagraph}[1]{\noindent \textbf{#1} ---}

\definecolor{Gray}{gray}{0.85}
\definecolor{GrayBorder}{gray}{0.65}
\definecolor{green_cylinder}{rgb}{0.0,0.40,0.0}
\definecolor{blue_cylinder}{rgb}{0.02,0.05,0.75}
\definecolor{way_point}{rgb}{0.56,0.0,1.0}
\newcolumntype{a}{>{\columncolor{GrayBorder}}c}

\title{\LARGE \bf
Self-supervised Pretraining and Finetuning for Monocular Depth and Visual Odometry
}
\author{Boris Chidlovskii and Leonid Antsfeld \\
Naver Labs Europe, Meylan, France \\
}

\begin{document}

\maketitle
\thispagestyle{empty}
\pagestyle{empty}

\begin{abstract}
For the task of simultaneous monocular depth and visual odometry estimation, we propose learning self-supervised transformer-based models in two steps. Our first step consists in a generic pretraining to learn 3D geometry, using cross-view completion objective (CroCo), followed by self-supervised finetuning on non-annotated videos. 
We show that our self-supervised models can reach state-of-the-art performance 'without bells and whistles' using standard components such as visual transformers, dense prediction transformers and adapters. We demonstrate the effectiveness of our proposed method by running evaluations on six benchmark datasets, both static and dynamic, indoor and outdoor, with synthetic and real images. For all datasets, our method outperforms state-of-the-art methods, in particular for depth prediction task.
\end{abstract}

\section{Introduction}
\label{sec:introduction}
\noindent

\begin{figure}[t]  \centering
    \includegraphics[width=0.97\linewidth]{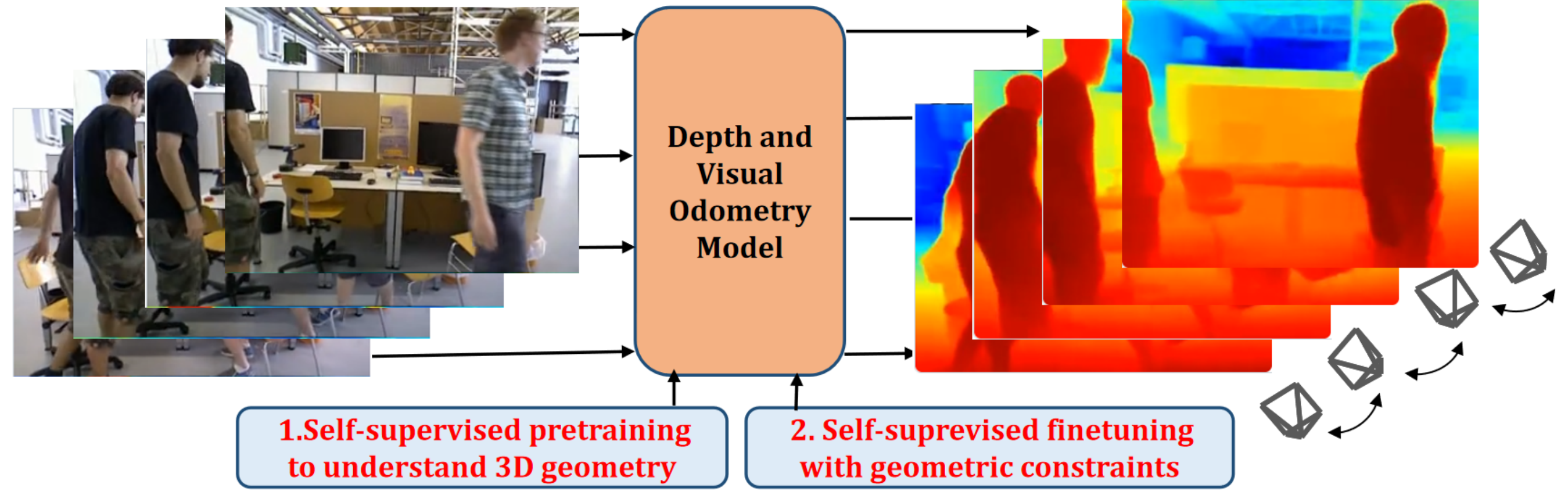} 
    \caption{\label{fig:dvo-def}
    Simultaneous depth and visual odometry estimation from a video, with two steps of model pretraining and finetuning.}
    \vspace*{-4mm}
\end{figure}

Visual odometry and depth estimation are critical for understanding the scene geometry and camera motion 
for tasks 
in robotics and autonomous driving. Supervised learning methods have been applied in many deep neural network frameworks and demonstrated outstanding results in both visual odometry and depth estimation~\cite{chang21transformerbased,piccinelli2023idisc,zhan20visualodometry}. However, supervised learning requires a significant amount of labeled data for training, 
thus the research interest moved towards the exploration of unsupervised learning frameworks that can learn  
scene depth and visual odometry (DVO) 
simultaneously (see Figure~\ref{fig:dvo-def}) 
without any ground truth annotations, using geometric constraints 
between adjacent frames.

Improving the network backbone of depth networks is 
a particularly effective way to gain in accuracy
~\cite{godard19digging,zhou17unsupervised}. 
Multiple sophisticated convolution-based backbones such as ResNet~\cite{godard19digging}, PackNet~\cite{guizilini203d}, HRNet~\cite{zhou21selfsuprevised}, HRDepth~\cite{lyu21hrdepth}, CADepth~\cite{yan22cadepth} 
have contributed to advances in the self-supervised monocular depth estimation task.  
Vision transformers (ViTs) have been deployed in supervised depth estimation as well~\cite{li2023depthformer,ranftl2022robust}, they were recently combined with convolutions in self-supervised monocular depth estimation~\cite{zhao22monovit}.

In this paper we argue that 
sophisticated backbone architectures are not crucial to the performance of DVO models, and they can be replaced with standard designs, such as generic transformers, 
thus benefiting from 
architecture unification and 
best training recipes, 
while retaining crucial properties of scalability, robustness and efficient transfer.
Our approach is inspired by recent advances in self-supervised pre-training to geometric vision tasks~\cite{gupta2023siamese,croco2022}. 
In particular,~\cite{croco2022} proposed the pretext task of {\it cross-view completion} (CroCo), a variant of masked image modelling (MiM) where a partially masked input image is reconstructed given visible patches and an additional view of the same scene. This pretraining objective is well suited to geometric downstream tasks such as optical flow and stereo matching~\cite{weinzaepfel2023croco}.
The CroCo architecture consists of a ViT encoder to extract features for the non-masked tokens of the first image, as well as for the second image, and a ViT decoder 
to reconstruct the masked image. 

We argue that CroCo is better suited for the DVO task than the existing backbones~\cite{godard19digging,lyu21hrdepth,ranft21vision,sun2022scdepthv3} 
which commonly use ImageNet-based pretrained models. Due to the object-centric~\cite{purushwalkam2020demystifying} and the balanced~\cite{assran2022hidden} nature of ImageNet~\cite{imagenet}, these models 
provide semantically rich features and excel on classification tasks, but can fail to capture structural information 
critical for dense geometric tasks like depth estimation and visual odometry.  
Instead, used as pretraining task, CroCo can easily reach state-of-the-art results without sophisticated network architecture, by using vanilla ViT blocks and standard geometric constraints on image reconstruction. 

In this paper, we show how to efficiently finetune the CroCo pre-trained model 
on monocular depth and visual odometry, in self-supervised way.
This is, to our best knowledge, the first task where {\it model pre-training and fine-tuning are both self-supervised}, with supervision signals coming from image completion at pretraining step, and from the geometric constraints at finetuning step. 

Driven by the concept of foundation models, a large scale pretraining model is trained on large heterogeneous datasets and targets a multitude of downstream tasks~\cite{yuan2021florence,croco2022}.
Such an approach represents an efficient alternative to curating a large dataset for one particular task as suggested in~\cite{ranftl2022robust,spencer2023kick}. In the case of DVO, we can benefit from the generic pretrained models oriented towards understanding 3D geometry of a scene, and finetune them on non-annotated videos to reach optimal downstream task performance. 

\begin{figure*}[t] \centering
    \includegraphics[width=0.9\linewidth]{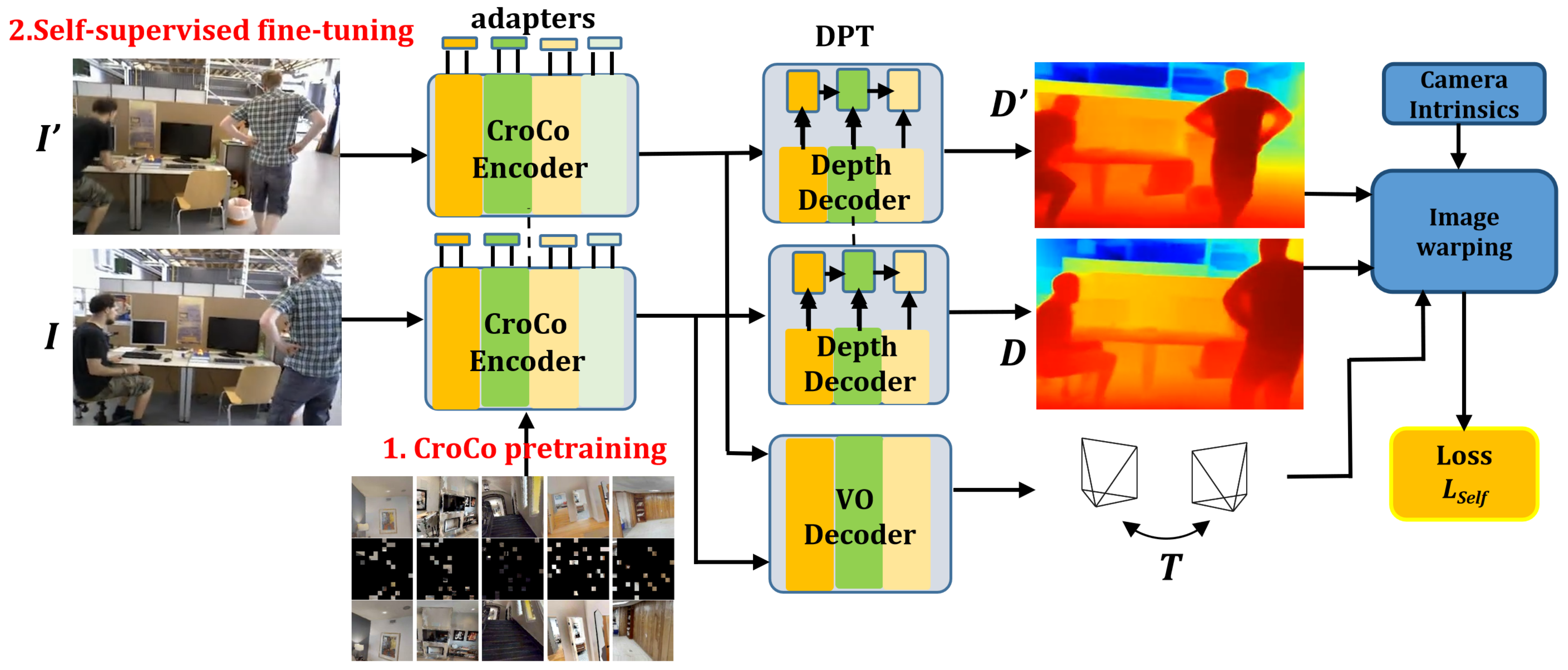} 
    \caption{\label{fig:DVO}
    \textbf{CroCo-DVO architecture}. \textcolor{red}{1}. Cross-view completion task is pretrained from a large amount of heterogeneous data. 
    \textcolor{red}{2}. At the finetuning step, given two consecutive frames ($I, I'$), we estimate their depth maps ($D, D'$) and relative pose $T$ using the network. 
    Then we compute the self-supervised loss $L_{Self}$ which is composed of  
    pixel-wise depth inconsistency between $D$ and $D'$, 
    geometric consistency loss  
    and a self-discovered mask 
    to handle dynamic objects 
    (see Section~\ref{ssec:losses}).  
}
 \vspace*{-4mm}
\end{figure*}

We reshape the pretrained model in a way that allows for self-supervised finetuning 
by sharing one encoder and targeting depth and visual odometry tasks with a proper decoder, by optimizing geometric consistency losses.

We claim the following contributions:
\begin{enumerate}
\item We propose a new approach for simultaneous depth and visual odometry estimation, where both pretraining and finetuning steps are self-supervised, with the learning signal coming from image cross completion and constraints on geometric consistency between adjacent frames.
\item We reshape the CroCo backbone for self-supervised finetuning to deal with the DVO tasks, 
and show how to enrich the backbone with recent (and already) standard extensions, such as adapters~\cite{chen22adapformer} for efficient transfer, and DPT for depth estimation~\cite{ranft21vision}.

\item The effectiveness of our proposed method is demonstrated through experiments on six benchmark datasets (NYUv2, KITTI, Tum, Bonn, Ddad and Gibson). We show that the model outperforms the-state-of-the-art methods on indoor and outdoor, static and dynamic, real and synthetic scenes.
\end{enumerate}

\section{Prior Work}
\label{sec:sota}
\noindent

\myParagraph{Self-supervised learning} 
Self-supervised learning is a way to learn generalizable visual representations often using different pretext tasks for pre-training~\cite{JingPAMI21SelfSupervisedVisualFeatureLearningDeepSurvey}
Inspired by BERT~\cite{BERT} in NLP,
different masked modeling methods (MIM) have been adapted to computer vision an robot perception. 
MIM pre-training aims at reconstructing masked information from an input image either in the pixel~\cite{mae22,SimMIM} or feature space~\cite{msn,maskfeat}.
Overall, MIM models perform well on classification tasks. They have obtained some success on denser tasks such as object detection~\cite{mae22} 
and have been applied to robotic 
perception~\cite{maerobot,nava21state}
when pre-trained on related datasets. Recently, CroCo~\cite{croco2022} introduced the pretext task of asymmetric cross-view completion, where a second (non-masked) view of the same scene is added to MIM. This is well suited to geometric downstream tasks: to leverage the second view and improve reconstruction accuracy, the model has to implicitly be aware of the geometry of the scene. 

\myParagraph{Unsupervised depth and visual odometry}
SfM-Learner~\cite{zhou_unsupervised_2017} was first to replace the known stereo baseline with an additional network to regress VO between consecutive frames and to supervise the training only by a stream of monocular images.  
Later, the explainability mask from SfM-Learner was replaced  with uncertainty~\cite{klodt18supervising} 
allowing the network to ignore incorrect predictions. 
DDVO~\cite{wang18learning} 
introduced a differentiable DSO module 
to refine the VO network prediction. 
Monodepth2~\cite{godard19digging} introduced  
minimum reconstruction filtering to reduce occlusion artifacts, alongside static pixel automasking via the raw reconstruction loss. 
D3VO~\cite{yang20d3vo} additionally predicted affine brightness transformation parameters for each support frame.  
SC-SfM-Learners~\cite{bian21unsupervised} and Sc-Depth-v3~\cite{sun2022scdepthv3} proposed an end-to-end differentiable geometric consistency constraint by synthesizing the support depth view. They included a variant of the absolute relative loss 
additionally used as auto-masking for the reconstruction loss. 
Alternatively, the depth models can be distilled~\cite{liu23self_supervised} and depth values can be discretized, 
by implicit partitioning the scene into a set of concepts in IDisc~\cite{piccinelli2023idisc},
or by capturing an input image as a set of orthogonal vertical planes and ground planes in PlaneDepth~\cite{wang23planedepth}.

\myParagraph{Network architectures} 
Multiple architectures used as backbone showed a significant impact on the performance of monocular depth estimation. First, GeoNet~\cite{yin_geonet_2018} 
replaced the VGG encoder with a ResNet. PackNet~\cite{guizilini203d} counts on 3D convolutions to learn detail-preserving compression and decompression of features. HRDepth~\cite{lyu21hrdepth} worked on the features decoding by implementing an attention module for multi-scale feature fusion.   
CADepth~\cite{yan22cadepth} extracts the long-range relationships between features via a channel-wise attention-based network to aggregate discriminated features in channel dimensions. 
Several recent works tackled supervised monocular depth estimation by using Transformer architectures~\cite{li2023depthformer,ranftl2022robust}. 
ViTs were also combined with convolutions in self-supervised monocular depth estimation~\cite{zhao22monovit}.


\section{Architecture}
\label{sec:architecture}

\subsection{Cross-view completion} 
\label{ssec:croco}
\noindent
\textit{Cross-View Completion} (CroCo)~\cite{croco2022} is a generic pretraining task trained on large heterogeneous dataset and which enables a network to perceive low-level geometric cues highly relevant to vision downstream tasks. It is an extension of masked image modeling~\cite{mae22}
processing pairs of images $(I,I')$, which correspond to two different views of the same scene with important overlap. In more details, the images are split into 
non-overlapping patches.  
The first input image $I$ is partially masked, 
the masked patches are discarded, and the remaining ones are fed to an image encoder, which is implemented using a ViT backbone. All patches from the second image are encoded 
using the same encoder with shared weights. The latent token representations output by the encoder from both images   
are then fed to a decoder whose goal is to predict the appearance of masked patches. It uses a series of transformer decoder blocks comprising cross-attention layers. This allows non-masked tokens from the first image to attend tokens from the reference image, thus enabling cross-view comparison and reasoning. The model is trained using a pixel reconstruction loss over all masked patches, similar to MAE~\cite{assran2022masked}. 
The CroCo pretext task has been shown to be applicable to different 
downstream vision problems; competitive performances were shown for the tasks of supervised monocular depth estimation~\cite{croco2022}, optical flow and stereo matching~\cite{weinzaepfel2023croco}.

\myParagraph{CroCo-DVO}
We now adapt CroCo pretrained model to self-supervised finetuning for DVO tasks. 
Our architecture includes two branches, for depth and visual odometry, which share the same encoder (see Figure~\ref{fig:DVO}). 
When finetuning the CroCo model, a pair of images ($I,I^{'}$) are fed to the encoder, then two decoders process the tokens of both images. For depth, to output a pixel-wise prediction, we initially rely on ViT layers with a linear MLP head per token, in order to output a dense depth map. The depth decoder consists in a series of transformer decoder blocks (self-attention among token features from the first frame, cross-attention with the token features from the second frame, and an MLP).
Finally, the second, VO decoder outputs the 6-DoF relative pose, it is designed as a lightweight MLP after patch-wise concatenation.

\myParagraph{DPT} 
As a powerful alternative to ViT depth decoder for dense tasks prediction, we propose to use the Dense Prediction Transformer (DPT)~\cite{ranft21vision}, which adapts the standard up convolutions and fusions from multiple layers 
to vision transformers. 
This allows to combine features from different blocks by reshaping them to different resolutions and fusing them with convolutional layers. In practice, the DPT uses the features from four blocks, regularly spread 
over the decoder depth, starting from the last block.  
In total, using transformers in DPT allows to make more detailed and globally consistent predictions 
~\cite{ranft21vision}.

\myParagraph{Adapters}
\label{ssec:adapters}
\begin{figure}[t]  \centering
    \includegraphics[width=0.9\linewidth]{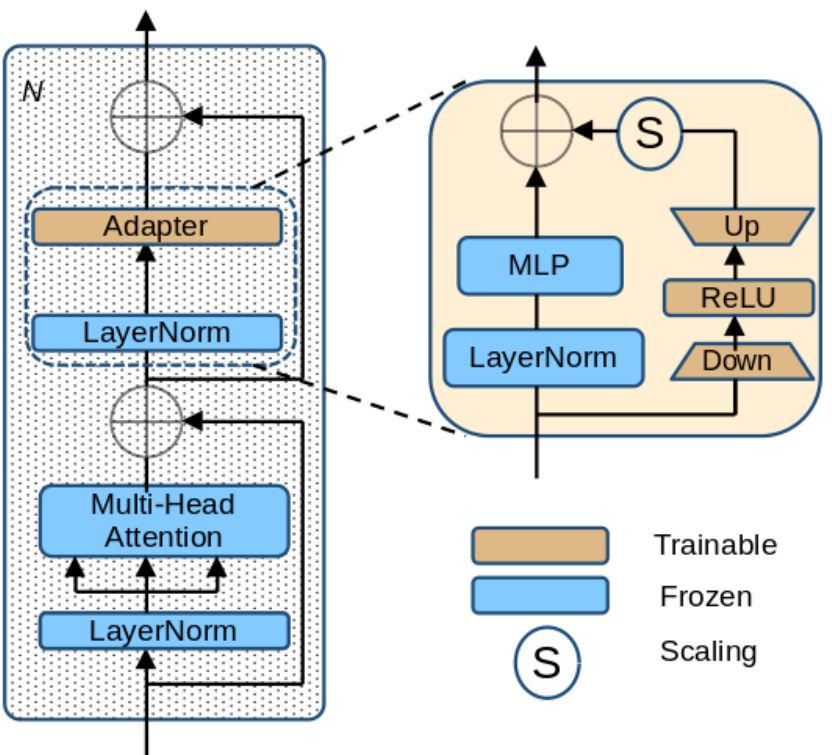} 
    \caption{\label{fig:adapters}
    \textbf{Adapters}: Training adapters while freezing the main backbone.
    AdaptFormer~\cite{chen22adapformer} replaces the MLP block in the transformer encoder with AdaptMLP, which is consisted of two sub-branches. The MLP layer in the left branch, identical to the original network, is frozen, right branch introduce a lightweight module for task-specific finetuning.}
\end{figure}
We also extend the CroCo-DVO backbone with adapters 
~\cite{chen22adapformer} which were designed for efficiently transfer large pretrained ViT-based models to downstream tasks. AdaptFormer attains strong transfer learning abilities by only finetuning a small number of extra parameters. 
The adapter architecture 
is shown in Figure~\ref{fig:adapters}. 
Compared to the full finetuning, AdaptFormer keeps the model parameters frozen and trains only additional residual network for each layer. In detail, they replace the MLP block in the transformer encoder with AdaptMLP, which consists of two sub-branches. The MLP layer in the left branch is identical to the original network, while the right branch is an additionally introduced lightweight module for task-specific finetuning. Speciﬁcally, the right branch is designed to be a bottleneck structure for limiting the number of parameters purpose, which includes a down-projection layer with parameters $W_{down} \in R^{d\times \hat d}$, an up-projection layer with parameters $W_{up} \in R^{d\times \hat d}$, where ${\hat d}$ is the bottleneck middle dimension and satisﬁes $\hat d << d$. 
In CroCo backbone, $d$ is 768 for encoder and 512 for decoder, and value of $\hat d$ is much smaller, usually 16 or 32.
The newly added parameters (right branch in Figure~\ref{fig:adapters}) are updated on the finetuning data with the self-supervised losses described in the following section.

\subsection{Self-supervised fine-tuning} 
\label{ssec:losses}
During the self-supervised finetuning step, depth estimation and visual odometry branches are simultaneously finetuned on a number of monocular videos. Given a pair of consecutive frames ($I, I'$) 
from a training video, we predict their depths $D$,$D'$ by using the depth decoder and estimate their relative 6-DoF camera pose $T$ by using the odometry decoder. 
%
With the obtained depth and relative camera pose, we follow  \cite{bian22csdepthv2,sun2022scdepthv3} in computing the warped ${\hat D}^{'}$ by transforming $D$ to 3D space (with known intrinsics) and projecting it to $I'$ using pose $T$. The inconsistency between warped ${\hat D}^{'}$ and predicted $D'$ is used as geometric consistency loss $L_g$ to supervise the training of the network.
The warping between two images using the predicted depth and pose is followed by synthesizing ${\hat I}^{'}$ by warping  
$I'$ via bi-linear interpolation. We penalize the color inconsistencies between $I$ and ${\hat I}^{'}$, and we constrain the geometry consistency between $D$ and $D'$, which backpropagates the gradients to the network. 

The first, {\it geometry consistency} loss $L_g$ aims 
to force the predicted depths $D$ and $D'$ to be consistent with each other in 3D space,
\vspace*{-3mm}
\begin{equation}
L_g = \frac{1}{|{\cal V}|} \sum_{p \in {\cal V}} D_{diff}(p),
\label{eq:lg}
 \vspace*{-2mm}
\end{equation}
where $\cal V$ is a set of valid points successfully projected 
from $I$ to $I'$, $D_{diff}$ stands for the pixel-wise depth inconsistency between $D$ and $D'$,
$D_{diff}(p) = \frac{|{\hat D}^{'}(p) - D'(p)|} {{\hat D}^{'}(p) + D'(p)}$.
To mitigate the influence of moving objects and occlusions that create geometrical inconsistency across multiple views, we follow~\cite{bian22csdepthv2} in using the self-discovered mask $m_s = 1 - D_{diff}$ which assigns lower weights to dynamic objects and occlusions. 

The second, {\it photometric} loss $L_p$ constrains the warping between $I$ and $I'$ generated by depth $D$ and pose $T$,
\begin{equation}
\begin{tabular}{l}
$L_p = \frac{1}{|\cal V|} \sum_{p \in \cal V} (1-\lambda)||I(p)-{\hat I}^{'}(p)||_1 + \lambda \frac{1-SSIM_{T}(p)}{2}$,\\
\end{tabular}
\label{eq:ssim}
\end{equation}
where ${\hat I}^{'}$ is synthesized from $I$ by warping, and SSIM is a widely-used metric to measure image structural similarity, $\lambda$ is 0.85 by following~\cite{godard19digging}. Actually, instead of loss $L_p$ in Eq.(\ref{eq:ssim}), we use its weighted version $L_p^m$ that counts on the mask $m_s$ to down-weigh regions of moving objects and obstacles,
\vspace*{-2mm}
\begin{equation}
\begin{tabular}{l}
$L_p^m =\frac{1}{|\cal V|} \sum_{p \in \cal V} (m_s(p) \cdot L_p (p))$, \\
\end{tabular}
\label{eq:lpm}
\vspace*{-2mm}
\end{equation}

The third, {\it edge-aware smoothness} loss regularizes the predicted depth map:
\vspace*{-2mm}
\begin{equation}
L_s = \sum_p (e^{-\Delta I(p)} \cdot \Delta D(p))^2, 
\label{eq:edge}
\vspace*{-2mm}
\end{equation}
where $\Delta$ is the first derivative along spatial directions, which guides smoothness by image edges.

The overall objective function is a weighted sum of the three loss terms presented above
\vspace*{-2mm}
\begin{equation}
L_{Self} = L_p^m + \beta L_g + \gamma L_s. 
\label{eq:total}
\vspace*{-2mm}
\end{equation}
In practice, we use $\beta$ = 0.5 and $\gamma$ = 0.1.  


\subsection{Implementation details}
\label{ssec:implementation}

We build on the publicly available CroCo code
\footnote{\scriptsize \url{https://github.com/naver/croco}.}. 
The CroCo-DVO encoder and decoder account for 12 and 8 ViT blocks, respectively. 
Four DPT layers are pulled from blocks 2, 4, 6, 8 of the CroCo decoder.
In both cases, the pixel-wise generation is terminated with 128 heads.
The VO decoder accounts for 2-layer MLP for the 6-DoF pose estimation.
We add adapters (of dimension $d'=32$) to both encoder and decoder layers.

\section{Experimental Results}
\label{sec:exp}
\subsection{Datasets and Metrics} 
We intensively evaluate our approach on a set of six benchmark datasets, which include indoor and outdoor, static and dynamic scenes, synthetically generated and real images. 
Outdoor datasets include KITTI~\cite{geiger13kitti} and Ddad~\cite{guizilini203d}; Indoor datasets include NYUv2~\cite{silberman12nyu}, Bonn~\cite{palazzo19refusion}, Tum~\cite{sturm12benchmark} and Gibson~\cite{igibson21} datasets. KITTI, NYUv2 and Gibson are mostly static datasets, while Ddad, Tum and Bonn datasets contain people or other dynamic objects. Gibson dataset includes synthetically generated images, others are the real image datasets. 
All the self-supervised methods are finetuned and tested on each dataset individually for a fair comparison with the state-of-the-art methods.

{\bf KITTI}. The dataset provides driving videos in urban scenes; it 
is the most widely-used dataset in self-supervised monocular depth estimation problems. 
We use Eigen’s split~\cite{eigen2015_predict} with 697 images for testing and the remaining video sequences for training. Depth ranges are capped at 80 meters, and images are resized to the resolution of 832 $\times$ 256 for training networks. 

{\bf NYUv2}. The dataset is widely-used in the computer vision community, it provides a large collection of indoor videos. There are 654 testing images of static scenes for depth evaluation, the remaining videos with 28,565 images are used for training. Images are resized to the resolution of $320 \times 256$. 

{\bf Ddad}. The dataset contains 200 driving videos captured in urban scenes.  
The standard split includes 150 training scenes (12,650 images) and 50 validation scenes (3,950 images) which are used for evaluation. 
Depth ranges are capped to at most 200 meters, and images are resized to the resolution of 640 $\times$ 384 for training the network.

{\bf Bonn}. The dataset contains 26 dynamic indoor videos.  
4 video sequences with fast-moving people (1785 images) are used for testing, and with the remaining videos for training. Depth ranges are capped at 10 meters, and images are resized to the resolution of 320 $\times$ 256. 

{\bf Tum}. . 
This collection consists of 11 sequences. 
Two sequences that contain moving people (1375 images) are used for testing. The remaining 9 dynamic videos are used for training, and images are resized to the the image resolution of 320 $\times$ 256. 

{\bf Gibson}. 
The Gibson dataset~\cite{xia2018gibson} is extensively used in indoor navigation tasks~\cite{chaplot2020object,PONI2022}, where the navigation goal is given by a point on a map, an image or an instance of object class and trajectories are traced by an expert.
We use the Gibson train split (72 scenes) for finetuning,  
and Gibson-val (14 scenes) to evaluate performance; image resolution is set to 320 $\times$ 256. 
 
We start from the pretrained CroCo model and finetune it for 50 epochs for NYUv2, Tum and Bonn datasets using AdamW optimizer and batch size of 12, 8, 8, respectively; the number of epochs is 25 for KITTI, Gibson and Ddad datasets, all with batch size of 4. The initial learning rate 
is $10^{-4}$ for all datasets 
except for KITTI where the initial learning rate is $10^{-5}$.

\myParagraph{Evaluation Metrics}
We use standard depth evaluation metrics~\cite{bian21unsupervised,zhou_unsupervised_2017} that include a) mean absolute relative error (AbsRel), b) root mean squared error (RMSE), 
and c)-e) the accuracy under threshold ($\delta_i < 1.25^i, i=1,2,3$).

Besides, we follow~\cite{bian21unsupervised,sun2022scdepthv3} in rescaling the predicted depth maps to match the 
median of the ground truth for evaluation, 
$s= median(D_{gt})/median(D_{pred}$).

\begin{table}[h] \centering
\setlength{\tabcolsep}{3pt}
\begin{tabular}{l|ccccc}
\specialrule{1pt}{0.5pt}{0.5pt} 
\toprule 
\rowcolor{GrayBorder}
Method & AbsRel $\downarrow$ & RMSE $\downarrow$ & $\delta_1 \uparrow$ & $\delta_2 \uparrow$ & $\delta_3 \uparrow$ \\ 
\hline
PackNet\cite{guizilini203d}& 0.109  & 4.696 & 0.884 & 0.961 & 0.981 \\ 
SC-SfM-Learners\cite{bian22csdepthv2}& 0.118 & 4.803 &0.866 &0.958 &0.981 \\
Sc-Depth-v3\cite{sun2022scdepthv3}& 0.118 & 4.709 &0.864 &0.960 &0.984 \\
CADepth-Net\cite{yan22cadepth} & 0.105 & 4.535 & 0.892 &0.964 &0.983 \\ 
DiffNet\cite{zhou21selfsuprevised}& 0.102 & 4.483 & 0.896 &0.965 &0.983 \\ 
MonoViT\cite{zhao22monovit}  & 0.099 & 4.372 & {\bf 0.900} &0.967 &0.984 \\ 
\hline
CroCo-DVO                  & 0.106 & 4.346 & 0.882 &0.966 &0.986 \\ 
CroCo-DVO+Ad               & 0.104 & 4.314 & 0.886 &0.967 &0.986 \\ 
CroCo-DVO+DPT              & {\bf 0.098} & {\bf 4.210} & {\bf 0.900} &{\bf 0.969} &{\bf 0.986} \\ 
\hline  
\end{tabular}
\caption{Self-supervised depth estimation on KITTI dataset.}
\vspace*{-6mm}
\label{tab:kitti}
\end{table}

\begin{table}[h] \centering
\setlength{\tabcolsep}{4pt}
\begin{tabular}{l|ccccc}
\specialrule{1pt}{0.5pt}{0.2pt}
\rowcolor{GrayBorder}
Method & AbsRel $\downarrow$ & RMSE $\downarrow$ & $\delta_1 \uparrow$  & $\delta_2 \uparrow$ & $\delta_3 \uparrow$ \\ \hline
Monodepth2~\cite{godard19digging}        & 0.169 & 0.614 & 0.745 & 0.946 & 0.987 \\
MonoIndoor~\cite{li2021monoindoortowards}& 0.134 & 0.526 & 0.823 & 0.958 & 0.989 \\
SC-SfM-Learners~\cite{bian22csdepthv2}   & 0.138 & 0.532 & 0.796 & 0.948 & 0.986 \\
Sc-Depth-v3~\cite{sun2022scdepthv3}      & 0.123 & 0.486 & 0.848 & 0.963 & 0.991 \\
\hline
CroCo-DVO             & 0.106 & 0.410 & 0.885 & 0.970 & 0.994 \\  
CroCo-DVO+Ad          & 0.101 & 0.398 & 0.892 & 0.976 & {\bf 0.995}\\ 
CroCo-DVO+DPT+Ad      &{\bf 0.095}&{\bf 0.377}&{\bf 0.905}&{\bf 0.980}&{\bf 0.995} \\  
\hline
\end{tabular}
\caption{Self-supervised depth estimation on NYUv2 dataset.}
\vspace*{-6mm}
\label{tab:nyu}
\end{table}
\begin{table}[bh] \centering
\setlength{\tabcolsep}{4pt}
\begin{tabular}{l|ccccc}
\specialrule{1pt}{0.5pt}{0.2pt}
\rowcolor{GrayBorder}
Method & AbsRel $\downarrow$ & RMSE $\downarrow$ & $\delta_1 \uparrow$  & $\delta_2 \uparrow$ & $\delta_3 \uparrow$ \\ 
\hline
Monodepth2~\cite{godard19digging} &0.312&1.408&0.474 &0.793 &0.905\\ 
SC-SfM-Learners~\cite{bian22csdepthv2} & 0.223 & 0.282 & 0.643 & 0.862 & 0.932 \\ 
Sc-depth-v3~\cite{sun2022scdepthv3}& {\bf 0.163} & 0.265 & 0.797 & 0.882 & 0.937 \\
\hline
CroCo-DVO                 &0.180 &     0.259 & 0.773 & {\bf 0.906} & 0.972  \\ 
CroCo-DVO+Ad              &0.182 &     0.252 &{\bf 0.752} & 0.903 & {\bf 0.978}  \\
CroCo-DVO+DPT+Ad          &0.180 &{\bf 0.251}& 0.756 & 0.905 & {\bf 0.978}  \\
\hline
\end{tabular}
\caption{Self-supervised depth estimation on Tum dataset.}
\label{tab:tum}
\vspace*{-8mm}
\end{table}

\begin{table}[h] \centering
\setlength{\tabcolsep}{4pt}
\begin{tabular}{l|ccccc}
\specialrule{1pt}{0.5pt}{0.2pt}
\rowcolor{GrayBorder}
Method & AbsRel $\downarrow$ & RMSE $\downarrow$ & $\delta_1 \uparrow$  & $\delta_2 \uparrow$ & $\delta_3 \uparrow$ \\ 
\hline
MonoDepth2~\cite{godard19digging}  & 0.565 & 2.337 & 0.352 &0.591 &0.728 \\ %
SC-SfM-Learners~\cite{bian22csdepthv2}& 0.211 & 0.619 & 0.714 &0.873 &0.936 \\
Sc-depth-v3~\cite{sun2022scdepthv3}& 0.126 & 0.379 & 0.889 &0.961 &0.980 \\ 
\hline
CroCo-DVO                          & 0.117 & 0.366 & 0.882 &0.971 &0.987 \\ 
CroCo-DVO+Ad                       &0.115  & 0.356 & 0.886 &0.972 &{\bf 0.989} \\ 
CroCo-DVO+DPT+Ad                   &{\bf 0.113}&{\bf 0.343}&{\bf 0.901}&{\bf 0.973} &0.988 \\ 
\hline
\end{tabular}
\caption{Self-supervised depth estimation on Bonn dataset.}
\label{tab:bonn}
\vspace*{-8mm}
\end{table}

\begin{table}[h] \centering
\setlength{\tabcolsep}{3pt}
\begin{tabular}{l|ccccc}
\specialrule{1pt}{0.5pt}{0.2pt}
\rowcolor{GrayBorder}
Method & AbsRel $\downarrow$ & RMSE$\downarrow$ & $\delta_1 \uparrow$ &$\delta_2 \uparrow$ &$\delta_3 \uparrow$ \\ 
\hline
Monodepth2~\cite{godard19digging}  &0.239 & 18.392 & 0.752 & 0.899 & 0.949 \\
SC-SfM-Learners~\cite{sun2022scdepthv3}&0.169 & 16.290 & 0.773 & 0.905 & 0.961 \\
Sc-depth-v3~\cite{sun2022scdepthv3}&{\bf 0.142}& 15.868 & 0.813 & 0.922 & 0.963 \\
\hline
CroCo-DVO                &0.156 & 14.509  & 0.789 & 0.929 & 0.971 \\
CroCo-DVO+Ad             &0.151 & 14.151  & 0.815 & 0.935 & 0.973 \\ 
CroCo-DVO+DPT            &{\bf 0.142} &{\bf 13.323}&{\bf 0.836} &{\bf 0.946}&{\bf 0.979} \\
\hline
\end{tabular}
\caption{Self-supervised depth estimation on Ddad dataset.}
\label{tab:ddad}
\vspace*{-6mm}
\end{table}

\subsection{Evaluation results}
\label{ssec:one}

In this section we evaluate various configurations of CroCo-DVO 
and extensions 
for the finetuning step. 
First, {\bf CroCo-DVO} is our baseline, it reuses and finetunes CroCo encoder and depth decoder with no change. Second, {\bf CroCo-DVO+Ad} adds and trains adapters in both encoder and decoder layers by freezing the large part of the network parameters. Third, {\bf CroCo-DVO+DPT} extends the baseline's depth decoder with four DPT layers for the accurate depth prediction.
Lastly, {\bf CroCo-DVO+DPT+Ad} combines the two extensions, adapters and DPT, in one network.

We use six datasets presented in the previous section to evaluate our models; the quantitative depth estimation results are reported in Tables~\ref{tab:kitti}-\ref{tab:gibson}. 
For all datasets except Gibson, we compare our method with the state-of-the-art methods, including Monodepth2~\cite{godard19digging}, PackNet~\cite{guizilini203d}, 
CADepth-Net\cite{yan22cadepth}, DiffNet\cite{zhou21selfsuprevised}, SC-SfM-Learners~\cite{bian22csdepthv2}, Sc-Depth-v3~\cite{sun2022scdepthv3} and MonoViT~\cite{zhao22monovit}. 
In each table we report the performance of our baseline and two other CroCo-DVO configurations. As results show, our method outperforms previous methods by a large margin in in four or all five metrics, 
thus demonstrating the efficacy of our CroCo-DVO architecture.

\begin{table}[ht] \centering
\setlength{\tabcolsep}{3pt}
\begin{tabular}{l|ccccc}
\specialrule{1pt}{0.5pt}{0.2pt}
\rowcolor{GrayBorder}
Method   & AbsRel $\downarrow$&RMSE$\downarrow$&$\delta_1 \uparrow$ &$\delta_2 \uparrow$ &$\delta_3 \uparrow$ \\ \hline
CroCo-DVO         &0.144 &0.307 &0.846 &0.952 &0.983 \\
CroCo-DVO+DPT     &0.125 &{\bf 0.246}&0.870 &{\bf 0.971}& {\bf 0.990}\\ 
CroCo-DVO+DPT+Ad  &{\bf 0.123} &0.250&{\bf 0.880}&0.961&0.985 \\
\hline
\end{tabular}
\caption{Self-supervised depth estimation on Gibson dataset.}
\label{tab:gibson}
\vspace*{-6mm}
\end{table}

\begin{figure*}[ht]  
\centering
\includegraphics[width=0.97\linewidth]{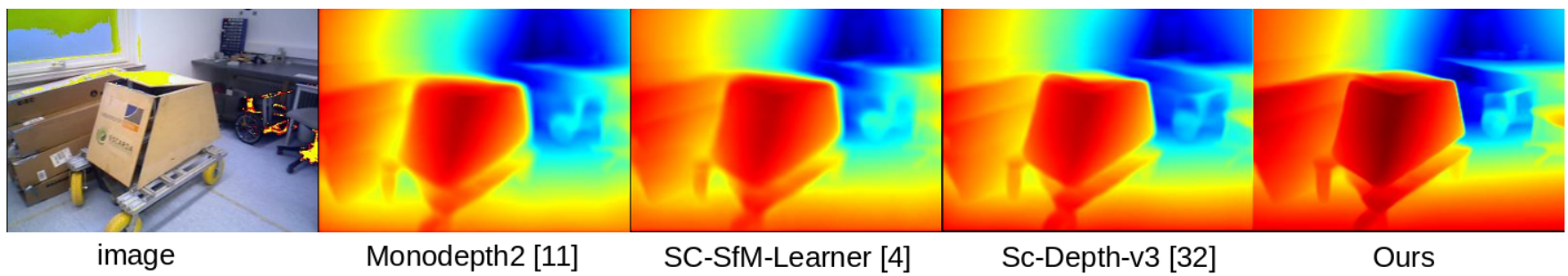} 
\caption{Qualitative depth estimation for the state-of-the-art and our methods.}
\label{fig:depth}
\vspace*{-2mm}
\end{figure*}

Self-supervised finetuning the CroCo backbone on DVO tasks already improves the state-of-the-art on two datasets, Bonn and NYUv2. It also shows competitive results on KITTI, Tum and Ddad datasets. Adapters and DPT plays a key role in the performance and improves the state-of-the-art. Adding extensions to the CroCo backbone reports the best results by all metric except AbsRel on Tum and Ddad datasets.
Our results is particularly important for KITTI and NYUv2 datasets, which constitute the most difficult and competitive benchmarks. 
 
The analysis shows that DPT systematically improves the depth prediction, in particular it makes blurry regions sharper. 
The advantage of using adapters is more subtle. It is very consistent on NYUv2, Tum, and Bonn datasets while yielding more limited benefits on Kitti and Ddad datasets. We hypothesize that outdoor driving scenes which compose these two datasets are not well represented in the large-scale dataset the CroCo backbone was pretrained on.
Training a small adapter network is insufficient to capture the specific street geometry. As result, CroCo-DVO-DPT that finetunes the entire network performs better. 
\begin{figure*}[ht]  
\centering
\includegraphics[width=0.97\linewidth]{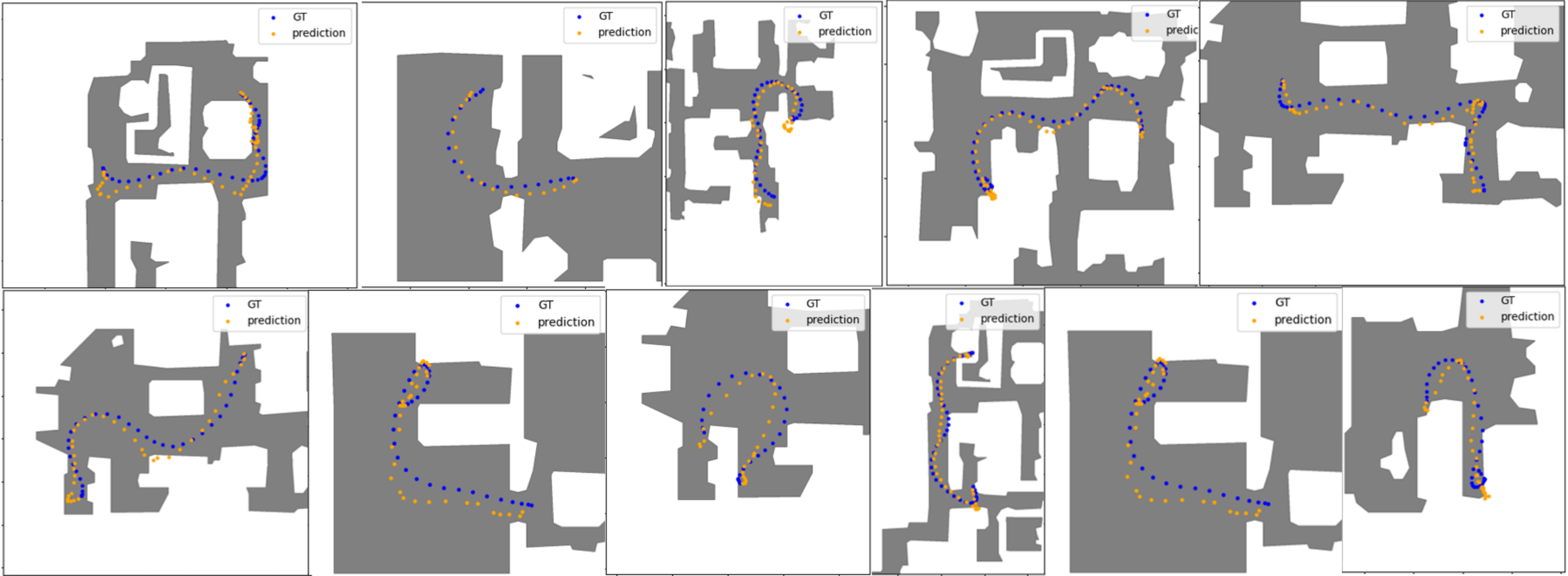} 
\caption{
\textbf{Visual odometry estimation: Estimated Gibson trajectories (orange) vs GT trajectories (blue).}}
\label{fig:traj}
\vspace*{-5mm}
\end{figure*}

\subsection{CroCo-DVO ablations}
\label{ssec:eval-adapter}
We now ablate components we described in Section~\ref{sec:architecture} to design the CroCo-DVO architecture. First, in Table~\ref{tab:adapter} we compare the CroCo-DVO baseline to CroCo-DVO+Ad with the frozen backbone and adapters dimension between 8 and 128, while finetuning on 
Bonn dataset. 
\begin{table}[ht]
 \centering
\setlength{\tabcolsep}{3pt}
\begin{tabular}{l|cccccc}
\specialrule{1pt}{0.5pt}{0.2pt}
\rowcolor{GrayBorder}
Metrics             & No ads& dim=8 & dim=16& dim=32& dim=64& dim=128 \\ \hline
$\delta_3 \uparrow$ & 0.987 & 0.988 & {\bf 0.989} & {\bf 0.989} & 0.988 & 0.983 \\
AbsRel $\downarrow$ & 0.117 & 0.120 & 0.116 & 0.116 & {\bf 0.115} & 0.130 \\
RMSE   $\downarrow$ & 0.366 & 0.359 & 0.358 & {\bf 0.356} & {\bf 0.356} & 0.374 \\
\hline
\end{tabular}
\caption{Ablation of the adapter dimensionality on Bonn dataset.} 
\label{tab:adapter}
\vspace*{-6mm}
\end{table}
The table reports the optimal values for AbsRel, RMSE and $\delta_3$ metrics when the adapter dimension is between 16 and 64. In dataset evaluations, the dimension of adapters is fixed to 32.

Second, we ablate more configurations of the CroCo backbone extended with adapters and DPT. 
Adapters are used in all layers by freezing the main backbone and training the small adapter network. 
Adding adapters increases the number of parameters by 2\% but reduces the finetuning time by 65\%. DPT layers, as presented in Section~\ref{fig:DVO}, are pulled from the depth decoder. In addition, we test a configuration where DPT is branched to the encoder, by pulling from blocks 3, 6, 9, 12, and skipping the decoder. 
Table~\ref{tab:ablation} presents six different configurations and reports their performance on Bonn dataset. As the table shows, contributions brought by the DPT and adapters are complementary. Used jointly, adapters and DPT with the depth decoder bring the most important improvement to the CroCo-DVO baseline. 

\begin{table}[ht] \centering
\setlength{\tabcolsep}{3pt}
\begin{tabular}{l|c|c|ccccc}
\specialrule{1pt}{0.5pt}{0.2pt}
\rowcolor{GrayBorder}
Method           &Ads & DPT&AbsRel$\downarrow$&RMSE$\downarrow$&$\delta_1$&$\delta_2$ &$\delta_3$ \\ 
{\bf DVO}       &         &      & 0.117  & 0.366 & 0.882 &0.971 &0.987 \\ 
{\bf DVO+Ad}    &  \check &      & 0.115  & 0.356 & 0.886 &0.972 &{\bf 0.989} \\  
DVO-DPT(E)      &         & Enc  & 0.117  & 0.362 & 0.886 &0.972 &0.987 \\ 
{\bf DVO+DPT}   &         & Dec  & 0.115  & 0.353 & 0.888 &0.971 &0.987 \\ 
DVO+DPT(E)+Ad   &\check   & Enc  & 0.115  & 0.357 & 0.886 &0.971 &0.988 \\ 
{\bf DVO+DPT+Ad}&\check   & Dec  &{\bf 0.113}&{\bf 0.343}&{\bf 0.901}&{\bf 0.973} &0.988 \\ 
\hline
\end{tabular}
\caption{Six CroCo-DVO configurations on Bonn dataset.}
\vspace*{-6mm}
\label{tab:ablation}
\end{table}
 
\subsection{Visual Odometry Evaluation}
\label{ssec:odometry}
Beyond the depth estimation, 
we use CroCo-DVO to estimate the visual odometry on the Gibson dataset, where the ground truth 6-DoF poses are available for all trajectories in the validation set. This allows us to 
evaluate the translation and rotation errors and visually compare the estimated and ground truth trajectories.
Taking into account that robots navigate on 2D-plane (which is not the case of other datasets) we project predicted trajectories on the xy-plane, and use the mean square error (MAE) to estimate the mean translation error (in meters) and the mean angular error is (in degrees) 
for all trajectories in the validation set (see Table~\ref{tab:vo-eval}). 
\begin{table}[ht] \centering
\setlength{\tabcolsep}{1pt}
\begin{tabular}{l|cc}
\specialrule{1pt}{0.5pt}{0.2pt}
\rowcolor{GrayBorder}
Method           & Translation Error(m) & Rotation Error($^{\circ}$) \\ \hline
CroCo-DVO        & 0.183             & {\bf 3.65}  \\
CroCo-DVO+Ad     & 0.181             & 3.71  \\
CroCo-DVO+DPT+Ad & {\bf 0.178}       & 3.68  \\
\hline
\end{tabular}
\caption{Translation and rotation errors for Gibson dataset.}
\label{tab:vo-eval}
\vspace{-4mm}
\end{table}
As the table shows, better depth estimation with extended CroCo-DVO versions is not necessarily translated in better odometry estimation. All three versions of CroCo-DVO report close values of translation (0.178-0.183 m) and rotation (3.65-3.71$^{\circ}$) errors.
Figure~\ref{fig:traj} completes this section; it proposes qualitative comparison between predictions and ground truth for a subset of  trajectories from Gibson validation set.

\section{Conclusion}
\label{sec:conclusion}
\noindent 
We proposed a ViT-based model for the self-supervised 
estimation of monocular depth and visual odometry. It benefits from the generic pretraining oriented towards understanding 3D geometry of any scene followed by self-supervised finetuning on non-annotated streams of images. 
We show that the model, standalone or completed with standard extensions like adapters or DPT, can easily reach the state-of-the-art performance 'without bells and whistles'. The effectiveness of our proposed method is demonstrated through experiments on six benchmark datasets, where our method outperforms the state-of-the-art methods in the depth estimation, both indoor and outdoor, static and dynamic, real and synthetic image scenes. 
\newpage 
\bibliographystyle{plain}
\bibliography{navigation,egomotion}

\end{document}